%% file: main.tex
\documentclass[10pt,twocolumn,letterpaper]{article}
\usepackage{cvpr}
\usepackage{booktabs}
\input{preamble}

\definecolor{cvprblue}{rgb}{0.21,0.49,0.74}
\usepackage[pagebackref,breaklinks,colorlinks,allcolors=cvprblue]{hyperref}
\def\confName{CVPR}
\def\confYear{2027}

\title{RePlan-VLA: Progress-Aware Replanning for Long-Horizon Vision-Language-Action Models}

\newcommand{\emaillink}[2]{\href{mailto:#1}{{\normalcolor #2}}}

\author{
{
Apoorva Vashisth$^{1,\ast}$ \quad
Manav Kulshrestha$^{1,*}$ \quad
Damon Conover$^2$
}\\[2pt]
{
Zachary Kingston$^{1,\dagger}$ \quad
Aniket Bera$^{1,\dagger}$
}\\[4pt]
$^1$Purdue University, West Lafayette, IN, USA\\
$^2$DEVCOM ARL, Adelphi, MD, USA\\[3pt]
{\ttfamily\small
\{%
\emaillink{vashista@purdue.edu}{vashista},%
\emaillink{mkulshre@purdue.edu}{mkulshre},%
\emaillink{zkingston@purdue.edu}{zkingston},%
\emaillink{aniketbera@purdue.edu}{aniketbera}%
\}@purdue.edu%
}\\
{\ttfamily\small
\emaillink{damon.m.conover.civ@army.mil}{damon.m.conover.civ@army.mil}
}
}

\usepackage{cuted}
\begin{document}
\maketitle

\begingroup
\renewcommand{\thefootnote}{$\ast$}
\footnotetext{Equal Contribution.}
\endgroup

\begingroup
\renewcommand{\thefootnote}{${\dagger}$}
\footnotetext{Equal advising.}
\endgroup

\input{sec/0_abstract}    
\input{sec/1_intro}
\input{sec/2_related_work}
\input{sec/3_our_approach}
\input{sec/4_Experiments}
\input{sec/5_Conclusion}
{
    \small
    \bibliographystyle{ieeenat_fullname}
    \bibliography{main}
}

\end{document}

%% file: preamble.tex
\usepackage{booktabs}
\usepackage{multirow}
\usepackage[table]{xcolor}
\usepackage[most]{tcolorbox}
\usepackage{bbm}
\usepackage{float}
\usepackage{graphicx}
\usepackage{listings}
\newtcolorbox{promptbox}[1]{
    enhanced,
    breakable,
    colback=black!2,
    colframe=black!35,
    colbacktitle=black!8,
    coltitle=black,
    fonttitle=\bfseries\small,
    title={#1},
    fontupper=\small,
    boxrule=0.4pt,
    arc=1.5mm,
    left=8pt,
    right=8pt,
    top=7pt,
    bottom=7pt,
    before skip=8pt,
    after skip=10pt
}

%% file: sec/0_abstract.tex
\begin{abstract}
Vision-Language-Action (VLA) models have emerged as a promising paradigm for general-purpose robotic manipulation, yet their performance often degrades on long-horizon tasks that require multiple interdependent manipulations. 
Instead of burdening the VLA policies with long-horizon reasoning, memory, action generation, progress tracking, and error recovery, we argue for leveraging their short-horizon manipulation capabilities within an execution-aware framework. 
To this end, we introduce \textnormal{\textbf{RePlan-VLA}}, a VLA-agnostic framework that translates a VLA's short-horizon capabilities into reliable long-horizon execution.
RePlan-VLA explicitly monitors the robot's motion for progress evaluation, and adaptively revises the future plan based on observed outcomes. 
We preserve the underlying VLA policy entirely, requiring no additional fine-tuning.
Experiments on RoboCasa365 and CALVIN benchmarks show an improvement of $21.9\%$ and $10.4\%$ in success rate over the next strongest baseline. 
We also validate our approach for real-world deployment on a Franka Research 3 robot.
\end{abstract}

%% file: sec/1_intro.tex
\section{Introduction}
\label{sec:introduction}

Vision-Language-Action (VLA) models~\cite{awomo2026awomo05, team2026xiaomi, intelligence2025pi_, nvidia2025gr00tn16, li2026trm, marmotlab2025falcon} have made significant progress toward developing generalist robotic policies capable of performing diverse manipulation tasks. 
Despite their performance on short-horizon tasks (e.g. `Pick up the straw and place it in the cup.'), these models often struggle with reliable execution of long-horizon instructions (e.g. ``Open the drawer, pick up the straw, navigate to the cup, and place the straw in the cup''). 
This can be attributed to their inability to coordinate simultaneous progress tracking and execution of the long-horizon tasks, allowing small errors to propagate and cause downstream failures.
Towards this end, we propose decoupling of long-horizon reasoning, action generation, progress tracking, and error recovery from a single-VLA policy to a VLA-agnostic framework, retaining the VLA policy only as a reliable executor of short-horizon tasks.

\begin{figure}[t]
    \centering
    \includegraphics[
        width=\linewidth,
        trim={180pt 0pt 180pt 0pt},
        clip
    ]{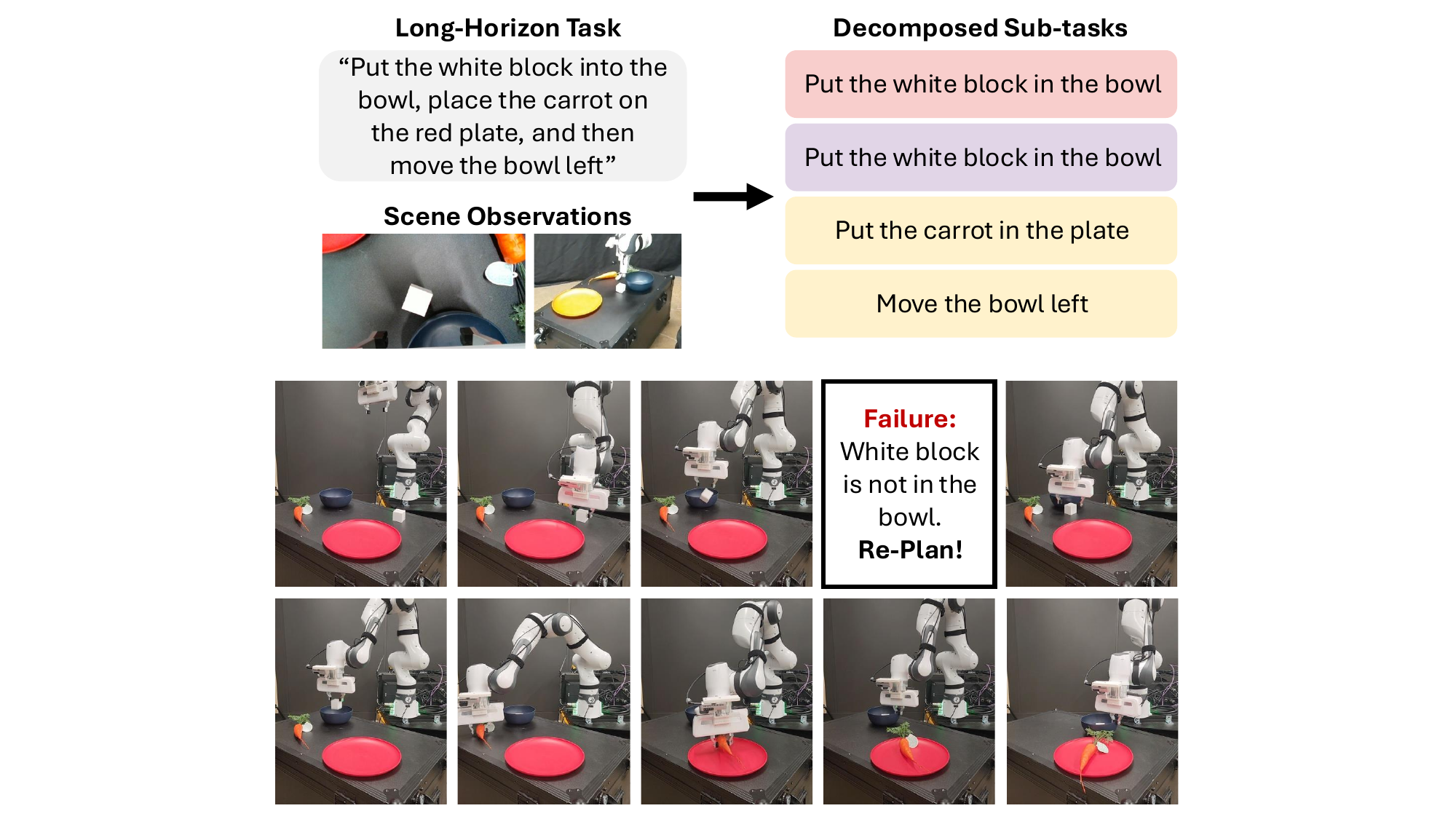}
    \caption{
    \textbf{RePlan-VLA: Execution-aware replanning for long-horizon robotic manipulation.}
    Given a long-horizon language instruction and scene observations, RePlan-VLA decomposes the task into executable short-horizon subtasks, monitors the robot's progress and detects discrepancies between intended and observed outcomes. 
    In this real-world rollout, the robot initially fails to place the white block into the bowl. 
    RePlan-VLA identifies the incomplete manipulation and replans to retry the failed subtask before proceeding with the remaining objectives. 
    }
    \label{fig:real_teaser}
    \vspace{-6mm}
\end{figure}

Recent approaches improve long-horizon manipulation through complementary mechanisms: leveraging future prediction~\cite{huang2026long, zhang2026foreact, chen2026unified}, persistent action memory~\cite{marrie2026reki, wu2024unleashing, huang2026long}, and action prediction~\cite{liu2026neural}.
Progress-aware approaches~\cite{zhang2026recurrent, Yan2026ProgressVLAPD, Li2026SentinelVLAAM}, either leverage progress estimates as monotonically increasing rewards for computationally intensive policy training, do not consider explicit failure diagnosis and recovery, or overlook the compositional structure and interdependencies of long-horizon tasks by treating them as monolithic execution problems.
Consequently, missed prerequisites and incorrectly assumed completions propagate through subsequent manipulations, motivating the need for an execution-grounded framework capable of handling task dependencies, progress monitoring, execution awareness, and failure-conditioned replanning.

\begin{figure*}[t]
    \centering
    \includegraphics[width=0.95\textwidth]{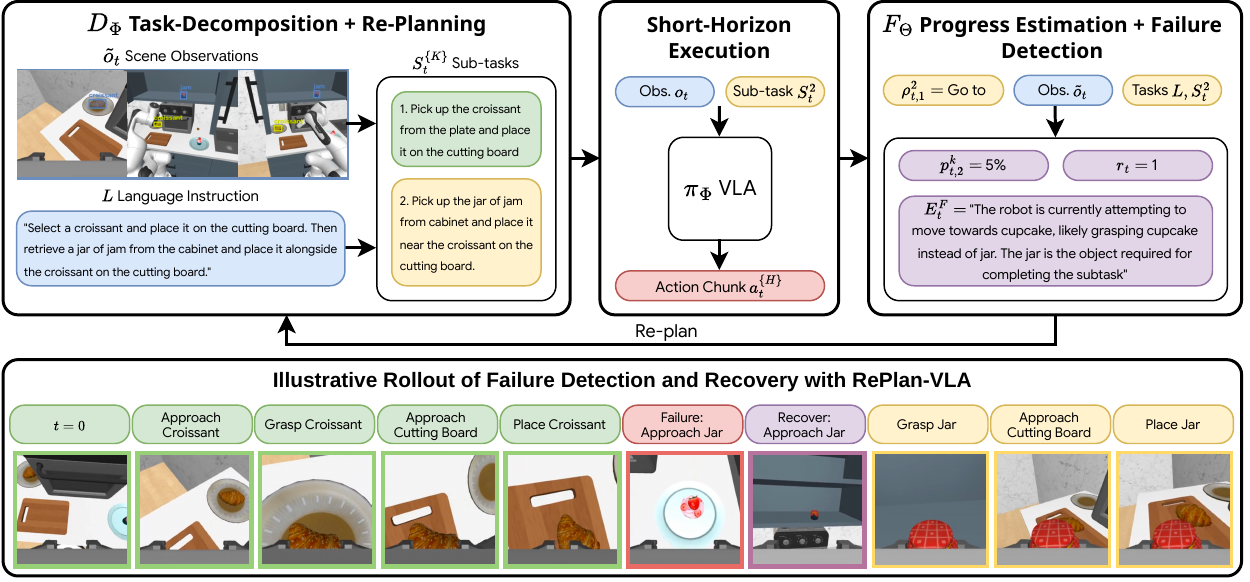}
    \caption{
    \textbf{Overview of RePlan-VLA.}
    Given a long-horizon language instruction and visually grounded scene observations, the planner decomposes the task into executable short-horizon subtasks, which are sequentially executed by a frozen VLA policy. A separate progress estimator monitors execution at the manipulation-primitive level, detects deviations from the intended subtask, and generates natural-language failure feedback. Upon failure, the planner uses the updated scene observations and diagnostic feedback to revise the remaining subtask sequence. The bottom panel illustrates this closed-loop process: after successfully placing a croissant on the cutting board, the robot approaches an incorrect object while retrieving a jam jar. RePlan-VLA detects the mismatch and replans, enabling the robot to recover and complete the remaining manipulation.}
    \label{fig:teaser}
    \vspace{-5mm}
\end{figure*}

To address these challenges, we propose a VLA-agnostic framework for long-horizon task execution. 
Our approach integrates language-based task decomposition, visual progress monitoring, and failure-conditioned re-planning into a closed-loop execution pipeline. 
We evaluate our approach on the RoboCasa365~\cite{nasiriany2026robocasa365} composite-task benchmark and a focused subset of challenging CALVIN's~\cite{mees2022calvin} long-horizon task sequences.
In summary, our main contributions are threefold:
\begin{itemize}
    \item We introduce \textbf{RePlan-VLA}, a VLA-agnostic framework that enables policies to reliably execute long-horizon tasks without requiring further VLA fine-tuning.

    \item We develop a fine-tuned progress-monitoring VLM that grounds sub-task progress estimation and failure diagnosis via visual observations to guide adaptive re-planning.

    \item We demonstrate the effectiveness of RePlan-VLA on challenging long-horizon manipulation benchmarks, improving average task success rate from $40.6\%$ to $62.5\%$ on RoboCasa365 composite tasks, and from $3.58/5$ to $4.10/5$ on CALVIN ABCD$\rightarrow$hardD.
\end{itemize}

%% file: sec/2_related_work.tex
\section{Related Work}
\label{sec:related}
\textbf{Vision-Language-Action Policies}. 
Early mobile manipulation approaches typically trained separate policies for individual tasks or narrowly defined sets of manipulation skills, limiting their ability to generalize to new instructions and environments.
RT-1~\cite{brohan2022rt} marked a shift toward generalist policies by learning multiple manipulation skills within a unified architecture, while RT-2~\cite{brohan2023rt} leveraged pretrained vision-language representations to improve semantic generalization. 
Building on these foundations, Octo~\cite{team2024octo} and OpenVLA~\cite{kim2024openvla} scaled training across diverse robot embodiments, enabling broader skill transfer and efficient adaptation. 
Subsequent approaches like $\pi_{0.5}$~\cite{intelligence2025pi_}, GR00T-N1.6~\cite{nvidia2025gr00tn16}, Xiaomi-Robotics-1~\cite{team2026xiaomi}, RLDX-1~\cite{kim2026rldx}, and PRTS~\cite{zhang2026prts} further advanced generalist manipulation through larger and more diverse training datasets, improved action-generation architectures, and refined training strategies. 
More recently, recognizing the limitations of purely reactive action prediction, World-Action Models such as Awomo-0.5~\cite{awomo2026awomo05}, PatchWAM~\cite{wang2026action}, and ABot-M0.5~\cite{chen2026abot} have explored coupling action generation with predictive visual world modeling to better capture environment dynamics and improve sequential decision-making.
Despite these advances, reliable long-horizon execution remains challenging, as these approaches largely emphasize policy-level capabilities without explicitly accounting for sub-task dependencies, execution-grounded progress tracking, or recovery from intermediate failures.

\noindent\textbf{Long-Horizon Planning.}
Several approaches extend VLAs beyond reactive action prediction through hierarchical subgoals (LoHo-VLA~\cite{yang2025lohovla}), temporal and motion reasoning (TRM-VLA~\cite{li2026trm}, HiF-VLA~\cite{lin2026hif}), and persistent memory (REKI~\cite{marrie2026reki}, Long-WAM~\cite{huang2026long}). 
Building on these advances, Foreact~\cite{zhang2026foreact}, GR-1~\cite{wu2024unleashing}, NIAF~\cite{liu2026neural}, and LoHo-Manip~\cite{liu2026lohomanip} incorporated future-state prediction and execution-aware planning. 
More recently, progress-aware methods (Progress-VLA~\cite{Yan2026ProgressVLAPD}, Sentinel-VLA~\cite{Li2026SentinelVLAAM}, SPR~\cite{Dai2026SeePR}, NeurVLA~\cite{liu2026neurvla}, R$^2$VLM~\cite{zhang2026recurrent}) have sought to monitor execution and guide recovery through progress estimates, often derived from temporal advancement in demonstrations or used as monotonic rewards for policy optimization. However, elapsed-time supervision may not reflect actual task completion, while monotonic progress signals can obscure regressions caused by execution errors. Consequently, maintaining an accurate representation of completed and failed sub-tasks, and translating these observations into corrective plans, remains challenging.

In contrast, our approach grounds progress estimation in the observed execution of individual manipulation primitives rather than elapsed time or monotonically increasing task-level progress. 
By explicitly tracking execution behavior, and using the diagnostic feedback to revise the remaining plan, our framework enables failure-aware recovery across interdependent sub-tasks without retraining or modifying the underlying VLA policy.

%% file: sec/3_our_approach.tex
\section{Our Approach}
\label{sec:approach}

\textbf{Overview}. 
Our framework, illustrated in~\cref{fig:teaser}, consists of three key components: 
(i) a few-shot \textbf{planner VLM} that decomposes long-horizon instructions into an ordered sequence of executable short-horizon subtasks and adaptively revises the plan during execution; 
(ii) a \textbf{progress estimator VLM} trained via supervised fine-tuning to monitor execution using temporal visual observations, task context, and sub-task completion history; and 
(iii) a \textbf{frozen VLA policy} that translates the active sub-task instruction into low-level robot control commands. 
To ensure compatibility with the underlying VLA, the planner generates sub-task instructions that align with the linguistic structure of instructions encountered during the VLA policy's training. 
During execution, the progress estimator periodically evaluates sub-task completion and advances the policy to the next sub-task upon successful execution. 
When progress stalls or execution deviates from the intended outcome, the estimator generates natural-language diagnostic feedback, allowing the planner to revise the remaining sub-tasks while preserving the history of completed actions. 
This closed-loop interaction between planning, progress estimation, and control enables RePlan-VLA to extend robust short-horizon policies to reliable long-horizon manipulation without re-training or modifying the underlying VLA.

\noindent\textbf{Problem Formulation.}
At time $t \in \mathbb Z_{\geqslant 0}$, given observations $o_t \in \mathcal O$ and a natural-language instruction $L \in \mathcal L$ defining the long-horizon task, we aim to generate and execute an action sequence $a^{\{N\}} \in \mathcal A^N$ to complete the task. %

Let $D_\Psi$ be a planner VLM that uses the current observation $o_t \in \mathcal O$ to decompose the instruction $L$ into an ordered sequence of short-horizon sub-tasks $S^{\{K\}}_t = (S_t^1, S_t^2, \dots, S_t^{K}) \in \mathcal S^+$ and associated sequences of manipulation primitives $P^{\{K\}}_t=(\rho_{t,1}^k, \ldots, \rho_{t,{M}}^k)$, where $K \in \mathbb Z_{>0}$ is the number of generated sub-tasks, $\rho_{t,j}^k \in \mathcal P$ and $M \in \mathbb Z_{>0}$ is the number of primitives for the sub-task $S_t^k$. 

During execution, a frozen VLA policy \(\pi_{\Phi}: \mathcal{O} \times \mathcal{S} \rightarrow \mathcal{A}^H\) takes the current observation \(o_t \in \mathcal{O}\) and the active subtask instruction \(S_t^k \in \mathcal{S}\) as inputs to predict an action chunk \(a_t^{\{H\}} \in \mathcal{A}^H\), where \(H\) denotes the number of actions in the predicted chunk.
To monitor execution, a progress estimator VLM $F_\Theta: \mathcal O^\tau \times \mathcal L \times \mathcal S \times \mathcal P \to [0, 1]\times \mathbb Z_2 \times \mathcal E$ periodically evaluates the active manipulation primitive $\rho^k_{t,j}$ in context of the task instructions $L, S^k_t$ using a temporally sampled observation history $o_t^{\{\tau\}} = (o_{t-\tau\delta}, \dots, o_{t-\delta}, o_t)$, 
where $\delta \in \mathbb{Z}_{>0}$ is the temporal sampling stride and $\tau$ is the number of sampled observations. Given this input, the progress estimator outputs the estimated progress $p^k_{t,j}$ of the active primitive $\rho^k_{t,j}$ pertaining to the current sub-task $S^k_t$, an indicator $r_t \in \mathbb Z_2$ for whether re-planning is required, and a natural-language summarization $E^F_t \in \mathcal E$ of the observed execution sequence.

A sub-task $S^k_t$ is considered complete when all its constituent primitives $P^k_t$ have been successfully completed, after which execution advances to the next sub-task $S^{k+1}_t$.

\noindent\textbf{Task Decomposition and Replanning}.
We employ a few-shot vision-language model (VLM) as the planner $D_\Psi$ to decompose the long-horizon instruction $L$ into an ordered sequence of executable short-horizon sub-tasks $S^{\{K\}}_t$ and their associated ordered sequences of manipulation primitives $P^{\{K\}}_t$. 
Our prompts are provided in the Appendix.
To facilitate scene-grounded reasoning, the planner $D_\Psi$ first identifies task-relevant objects from the long-horizon instruction $L$. 
These objects are subsequently localized using an off-the-shelf object detection model, and the observations $o_t$ are augmented with object labels and spatial annotations (see Appendix) to produce $\tilde{o}_t \in \mathcal O$. 
The annotated observations $\tilde o_t$ are then provided to the planner and produce $(S^{\{K\}}_t, P^{\{K\}}_t) \sim D_\Psi(\cdot\ |\ \tilde o_t, L)$. 
Conditioning on the annotated scene enables the planner to generate sub-tasks consistent with the observed environment (see ~\cref{tab:component_ablations}).

The set of manipulation primitives is given by $\mathcal P = \{\texttt{go\_to},\ \texttt{move\_to}\}$.
The \texttt{go\_to} primitive describes approaching and grasping a target object, whereas \texttt{move\_to} describes transporting the target toward its intended destination. 
For instance, the instruction ``Pick up the cherry and place it on the pie.'' is decomposed into $P_t^k = (\texttt{go\_to},\ \texttt{move\_to})$, corresponding to grasping the cherry and subsequently placing it on the pie.

An auxiliary LLM-based instruction verifier $V_{\psi}: \mathcal S \to \mathcal E$ independently evaluates each sub-task instruction $S^k_t$ against the acceptable language forms of the employed VLA $\pi_\Phi$. 
If an instruction is deemed invalid, the verifier $V_\psi$ provides a natural-language explanation $E^V_t$ which is used to re-prompt the planner to generate a revised sub-task instruction $S^k_t \leftarrow \tilde S^k_t$. 
This iterative verification process encourages the planner $D_\Psi$ to produce sub-task instructions compatible with the frozen VLA's ($\pi_\Phi$) language interface. 

During execution, the planner $D_\Psi$ is invoked again whenever the progress estimator $F_\Theta$ detects stalled progress or an unexpected execution outcome, via the indicator $r_t$. 
We maintain a structured execution history $\eta_t$ containing previously successful sub-task instructions.
This persistent history allows the planner $D_\Psi$ to preserve completed manipulations, avoid redundant actions, and generate recovery plans consistent with the robot's execution history.
Re-planning is conditioned on the original instruction $L$, the failed sub-task instruction $S^k_t$, the current annotated observation $\tilde{o}_t$, the execution history $\eta_t$, and the natural-language failure explanation $E^F_t$ provided by the progress estimator $F_\Theta$. 
The re-planning operation is formulated as $\tilde S_t^{\{K'\}}\sim D_\Psi(\cdot\ |\ \tilde o_t, L, \eta_t, S^k_t, E^F_t)$
where $\tilde S_t^{\{K'\}}$ represents the regenerated sequence of remaining sub-tasks and $K' \in \mathbb Z_{>0}$ is the number of regenerated sub-tasks. 
The planner can introduce missing prerequisites, account for changes in object locations, and generate corrective sub-tasks when observed execution outcomes deviate from the original plan, without repeating previously completed tasks.

\noindent\textbf{Visual Progress Estimation and Failure Detection}
We use a fine-tuned VLM as the progress estimator $F_\Theta$ to assess execution at the granularity of manipulation primitives  $\mathcal P$ rather than the entire long-horizon task $L$. 
Each sub-task's ($S^k_t$) associated ordered manipulation primitive sequence $P^k_t$ comprises \texttt{go\_to} and \texttt{move\_to} primitives, corresponding to approaching and grasping an object and subsequently transporting or manipulating it toward its destination.
This formulation enables fine-grained monitoring of sub-task completion and execution errors.

We fine-tune a pretrained VLM on short-horizon trajectories from RoboCasa365, CALVIN, and real-world demonstrations. 
Each trajectory is annotated with manipulation phases, and continuous progress labels are derived from spatial relationships between the gripper, target object, and destination.
A lightweight regression head maps the VLM's final hidden representation to a progress estimate $p^k_{t,j}\in[0,1]$, while the backbone is adapted using LoRA~\cite{hu2021lora}.
We train the progress estimator $F_\Theta$ using a Smooth L1 regression loss with ground-truth progress supervision.
This allows the estimator to generalize primitive-level progress monitoring to long-horizon tasks without requiring long-horizon training demonstrations. Data annotation, model architecture, and training details are provided in the Appendix.

During execution, $F_\Theta$ evaluates the active primitive $\rho^k_{t,j}$ via temporally sampled observations $o_t^{\{\tau\}} \in \mathcal{O}^\tau$, the original instruction $L$, and the active subtask $S_t^k$. 
Primitive completion is determined by thresholding the predicted value:$c^k_{t,j} = \mathbbm{1}[p^k_{t,j} \geqslant \kappa_{\rho^k_{t,j}}],$
where $\kappa_{\rho^k_{t,j}}$ denotes the primitive-specific completion threshold. 
The framework advances through the primitives until the active sub-task $S_t^k$ is complete. 
We provide analysis of the accuracy of our progress estimator in the Appendix.

Beyond numerical progress estimation, \(F_\Theta\) leverages its pretrained language reasoning capabilities, further adapted through LoRA fine-tuning for task-specific execution understanding, to identify stalled execution, incorrect object interactions, and deviations from the intended sub-task.
It generates a natural-language failure explanation $E^F_t$ and a binary re-planning decision $r_t\in\{0,1\}$ without separate failure-diagnosis supervision. When $r_t=1$, execution is interrupted and the diagnostic feedback $E^F_t$ is passed to the planner $D_\Psi$ to revise the remaining sub-tasks $S^{\{K\}}_t$. Thus, numerical progress estimates govern primitive completion, while language-based reasoning identifies failures requiring corrective re-planning. The effectiveness of this task-oriented adaptation is further supported by improvements over a zero-shot VLM baseline (see ~\cref{tab:component_ablations}).

\noindent\textbf{Vision-Language-Action Policy.}
RePlan-VLA employs a frozen VLA policy $\pi_\Phi$ as the low-level executor, translating short-horizon sub-task instructions into embodiment-specific control actions. At each timestep $t$, the policy $\pi_\Phi$ receives the active sub-task instruction $S_t^k$ and its native observations $o_t$, generating actions $S^k_t$. High-level planning and execution history are managed separately, allowing the VLA to operate solely on the active sub-task. Upon sub-task completion or failure-triggered re-planning, any unexecuted actions from the previous prediction chunk are discarded, and the policy receives the updated instruction.

RePlan-VLA is VLA-agnostic, supporting integration with language-conditioned policies without modifying their observation modalities, action representations, or action-chunking strategies.
Each policy is fine-tuned on the target benchmark following its recommended training protocol and remains frozen throughout RePlan-VLA execution, requiring no additional training or architectural modifications.
Further integration details are provided in the Appendix.

%% file: sec/4_Experiments.tex
\section{Experiments}
\label{sec:experiments}

\subsection{Setup}
\noindent\textbf{Benchmarks and Evaluation Metrics.}
We evaluate RePlan-VLA on RoboCasa365~\cite{nasiriany2026robocasa365} and CALVIN~\cite{mees2022calvin} benchmarks.
For RoboCasa365, we evaluate 16 tasks each from the target Composite-Seen and Composite-Unseen splits, with five episodes per task.
We report task success rate (SR).
For CALVIN, we follow the ABCD$\rightarrow$harD protocol and evaluate five-task instruction sequences on a challenging subset of 100 sequences selected based on low baseline performance to reduce saturation.
All methods use the same evaluation subset, and we report success rates at sequence lengths $k\in\{1,\ldots,5\}$ and the average number of consecutively completed tasks (Avg.\ Len.).
Further details on CALVIN subset selection are provided in the Appendix.

\noindent\textbf{Training Configuration.}
We train the progress estimator for $80,000$ steps on a single NVIDIA H100 GPU using BF16 mixed precision and AdamW optimization. 
We employ LoRA ($r=16$, $\alpha=32$) and a $1024$-dimensional progress prediction head, with an effective batch size of $8$ and gradient clipping at $1.0$. 
The head and LoRA parameters use learning rates of $10^{-4}$ and $10^{-5}$, respectively. 
Further training and data annotation details are provided in the Appendix.
All underlying VLAs are fine-tuned following their respective benchmark protocols and remain frozen during RePlan-VLA execution; additional VLA implementation details are provided in the Appendix. 

\noindent\textbf{Implementation Details}.
We implement RePlan-VLA using the Qwen3-VL-8B~\cite{bai2025qwen3vl} models as few-shot planner VLM and as base model for fine-tuned progress estimator VLM. 
We use Awomo-0.5I~\cite{awomo2026awomo05} as our frozen VLA executor. 
All inferences were performed locally. 
During execution, the progress estimator evaluates the active manipulation primitive every $\mu=50$ environment steps using a temporal history of five observations sampled at intervals of $s=10$ steps. 
We use completion thresholds of $0.8$ and $0.65$ for the \texttt{go\_to} and \texttt{move\_to} primitives, respectively. 
Additional implementation details are provided in the Appendix.

\noindent\textbf{Real-Robot Evaluation}.
We evaluate RePlan-VLA in a real-world manipulation setting using a Franka Research 3 robotic arm equipped with a two-finger gripper. 
Our setup is illustrated in~\cref{fig:real_teaser}. 
Inspired by CALVIN~\cite{mees2022calvin}, we design long-horizon composite tasks combining language-conditioned manipulation primitives into multi-step sequences. 
Additional details on real-robot data collection and implementation are provided in the Appendix.

\subsection{Baseline Comparison}

\subsubsection{Robocasa365}

\noindent\textbf{Baselines}.
We compare RePlan-VLA against nine representative approaches, spanning generalist robot policies, memory-augmented architectures, and long-horizon planning frameworks.
The generalist baselines include GR00T-N1.6~\cite{nvidia2025gr00tn16}, $\pi_{0.5}$~\cite{intelligence2025pi_}, RLDX-1~\cite{kim2026rldx}, Xiaomi-Robotics-1~\cite{team2026xiaomi}, and Awomo-0.5I~\cite{awomo2026awomo05}, representing diverse action-generation architectures and large-scale robot learning paradigms.
Foreact~\cite{zhang2026foreact} is particularly relevant as it introduces high-level planning and visual foresight without modifying the VLA architecture. 
We evaluate REKI-2B~\cite{marrie2026reki}, which incorporates persistent memory through register tokens, and Long-WAM~\cite{huang2026long}, which leverages extended visual histories and predictive world modeling.
Baseline implementation details are provided in the Appendix.
Additional performance analysis is reported in the Appendix.

\begin{table}[t]
\centering\small

\caption{
RoboCasa365 composite task benchmark results. We report task success rates (SR, \%) on composite-seen and composite-unseen task splits, along with their average. Higher is better; best results are highlighted in bold.
}
\label{tab:robocasa_main}

\setlength{\tabcolsep}{5pt}
\begin{tabular}{lccc}
\toprule
\multirow{2}{*}{Approach} & \multicolumn{3}{c}{Success Rate (\%) $\uparrow$} \\
\cmidrule(lr){2-4}
& Seen & Unseen & Avg. \\
\midrule

GR$00$T-N1.6~\cite{nvidia2025gr00tn16} & 10.0 & 0.00 & 5.00 \\

$\pi_{0.5}$~\cite{intelligence2025pi_} & 18.8 & 0.00 & 9.40 \\

RLDX-1~\cite{kim2026rldx} & 16.3 & 6.30 & 11.3 \\

Xiaomi Robotics-1~\cite{team2026xiaomi} & 36.3 & 27.5 & 31.9 \\

Awomo-0.5I~\cite{awomo2026awomo05} & \underline{55.0} & \underline{26.3} & \underline{40.6} \\

\midrule

Foreact~\cite{zhang2026foreact} & 6.30 & 0.00 & 3.15 \\

Long-WAM~\cite{huang2026long} & 15.8 & 6.10 & 10.9 \\

REKI-2B~\cite{marrie2026reki} & 32.0 & 14.1 & 23.1 \\

Long-WAM + GPT-6 Astra~\cite{huang2026long} & 38.8 & 35.0 & 36.9 \\

\midrule

\textbf{RePlan-VLA} & \textbf{68.8} & \textbf{56.3} & \textbf{62.5} \\ \bottomrule

\end{tabular}
\vspace{-5mm}
\end{table}

As shown in~\cref{tab:robocasa_main}, RePlan-VLA achieves the highest average success rate of $62.5\%$, outperforming Awomo-0.5I ($40.6\%$) by $21.9$ percentage points. 
General-purpose VLAs such as GR00T-N1.6, $\pi_{0.5}$, and RLDX-1 achieve only $5.0$--$11.3\%$, while Xiaomi Robotics-1 reaches $31.9\%$. 
Their limited performance suggests that strong short-horizon manipulation capabilities alone do not ensure reliable execution of multi-stage tasks. %
Long-horizon methods Foreact, Long-WAM, and REKI-2B attain $3.2\%$, $10.9\%$, and $23.1\%$, respectively, while Long-WAM augmented with GPT-6 Astra improves to $36.9\%$. 
These results suggest that temporal context and high-level planning are insufficient when the system does not attempt to reconcile intended actions with observed outcomes. 
Awomo-0.5I performs comparatively well on seen tasks ($55.0\%$) but drops to $26.3\%$ on unseen compositions, indicating difficulty generalizing across novel task sequences. 
In contrast, RePlan-VLA achieves $56.3\%$ on unseen tasks, exceeding the strongest unseen baseline by $21.3$ points. 
These findings are consistent with the benefit of explicitly tracking completed subtasks, diagnosing execution failures, and adapting subsequent instructions without modifying the underlying VLA.

\begin{table}[t]
\centering
\small
\caption{CALVIN ABCD$\rightarrow$harD 5-task sequence results.
Columns 1--5 denote success rates (\%) for completing that many
consecutive tasks; Avg. is mean completed tasks out of five.
Higher is better.}
\label{tab:calvin_main}
\setlength{\tabcolsep}{5pt}
\begin{tabular}{lcccccc}
\toprule
Approach & 1 & 2 & 3 & 4 & 5 & Avg. \\
\midrule
$\pi_{0.5}$~\cite{intelligence2025pi_} & 80.0 & 42.0 & 20.0 & 16.0 & 10.0 & 1.68 \\

FLOWER~\cite{reuss2025flower} & 94.0 & 68.0 & 44.0 & 42.0 & 38.0 & 2.86 \\

MDT~\cite{reuss2024multimodal} & \underline{96.0} & 72.0 & 48.0 & 46.0 & 40.0 & 3.02 \\

XR-0~\cite{Cai2026XiaomiRobotics0AO} & 90.0 & \underline{78.0} & \underline{66.0} & 64.0 & \textbf{60.0} & \underline{3.58} \\

MoDe~\cite{reuss2025efficient} & 86.0 & 62.0 & 46.0 & 36.0 & 30.0 & 2.60 \\
\midrule

ForeACT~\cite{zhang2026foreact} & 62.0 & 22.0 &  6.00 &  4.00 &  00.0 & 0.94 \\

GR-1~\cite{wu2024unleashing} & 80.0 & 64.0 & 46.0 & 30.0 & 24.0 & 2.44 \\

NIAF~\cite{liu2026neural} & \underline{96.0} & 76.0 & 44.0 & 34.0 & 28.0 & 2.78 \\

\midrule

\textbf{RePlan-VLA} & \textbf{100.0} & \textbf{90.0} &
\textbf{86.0} & \textbf{76.0} & \underline{58.0} &
\textbf{4.10} \\

\bottomrule
\end{tabular}
\vspace{-6mm}
\end{table}

\subsubsection{CALVIN ABCD$\rightarrow$harD}

\noindent\textbf{Baselines}.
We compare RePlan-VLA against $8$ representative approaches on the CALVIN long-horizon benchmark, spanning generalist VLAs, diffusion-based policies, and methods incorporating spatial or temporal reasoning. 
Generalist baselines include $\pi_{0.5}$~\cite{intelligence2025pi_} and Xiaomi-Robotics-0~\cite{Cai2026XiaomiRobotics0AO}, while FLOWER~\cite{reuss2025flower}, MDT~\cite{reuss2024multimodal}, and MoDE~\cite{reuss2025efficient} represent diffusion-based visuomotor policies. 
We additionally consider 
GR-1~\cite{wu2024unleashing}, which leverages video prediction for manipulation. 
Finally, Foreact~\cite{zhang2026foreact} and NIAF~\cite{liu2026neural} provide comparisons with approaches designed for long-horizon execution. 
All methods are evaluated on the same challenging subset of 50 CALVIN ABCD$\rightarrow$D sequences (dubbed harD), with implementation details provided in the Appendix

As shown in~\cref{tab:calvin_main}, RePlan-VLA achieves the highest average sequence length of $4.10/5$, surpassing the strongest baseline, XR-0 ($3.58$), by $0.52$ tasks. 
While $\pi_{0.5}$, FLOWER, MDT, and MoDe exhibit substantial performance degradation as sequence length increases, their results suggest that reliable individual manipulation skills do not necessarily translate into sustained multi-task execution. 
Foreact achieves only $0.94$ average tasks, indicating that anticipatory action generation alone may struggle on challenging task sequences. 
Similarly, GR-1 ($2.44$) leverages future-observation prediction, while NIAF ($2.78$) improves temporal action modeling, but neither explicitly diagnoses failed subtasks to revise subsequent execution. 
RePlan-VLA maintains $86\%$ success through three consecutive tasks and $76\%$ through four, highlighting the benefit of execution-grounded progress monitoring and corrective replanning. 
Notably, XR-0 slightly outperforms RePlan-VLA on complete five-task sequences ($60\%$ vs.\ $58\%$), suggesting that strong underlying manipulation capabilities remain important even with explicit replanning.
Additional performance analysis is reported in the Appendix.

\subsection{Component and Design Ablations}

\begin{figure*}[t]
    \centering

    \begin{subfigure}[t]{\textwidth}
        \centering
        \includegraphics[width=\linewidth]{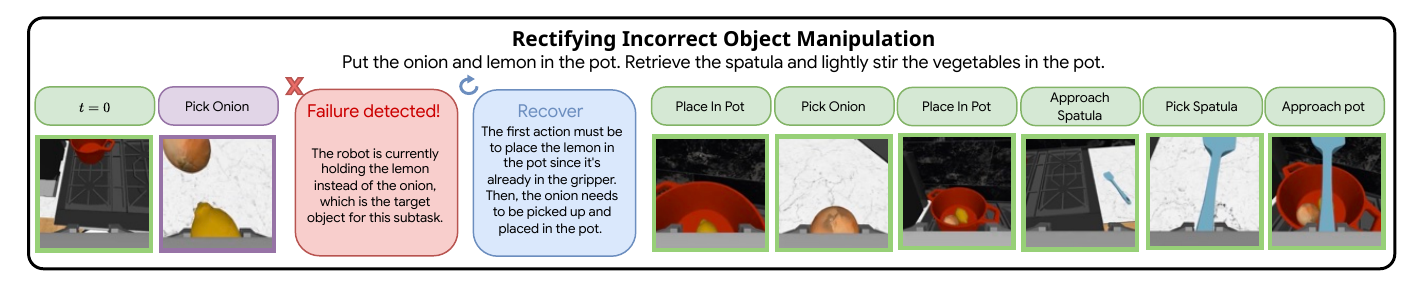}
        \caption{Recovering from an incorrect object grasp by placing the already-held lemon before returning to the intended onion manipulation.}
        \label{fig:incorrect_object}
    \end{subfigure}

    \vspace{0.15cm}

    \begin{subfigure}[t]{\textwidth}
        \centering
        \includegraphics[width=\linewidth]{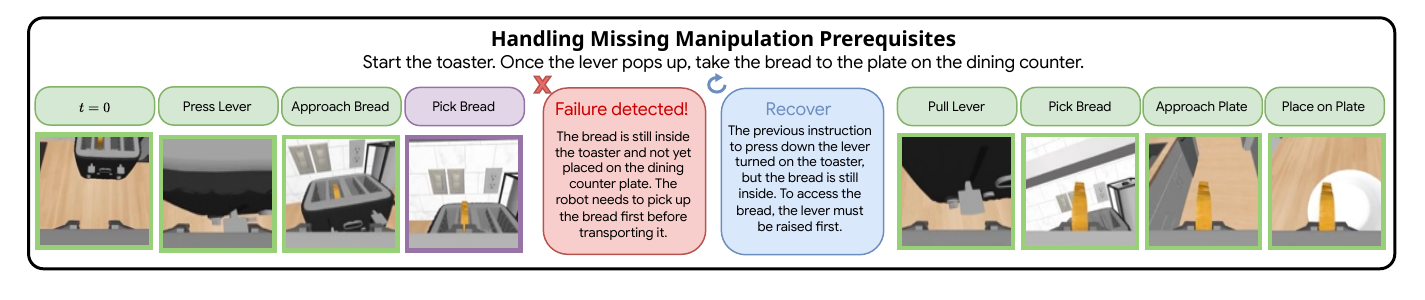}
        \caption{Restoring a missing prerequisite by raising the toaster lever to access the bread before transferring it to the plate.}
        \label{fig:missing_prerequisite}
    \end{subfigure}

    \vspace{0.15cm}

    \begin{subfigure}[t]{\textwidth}
        \centering
        \includegraphics[width=\linewidth]{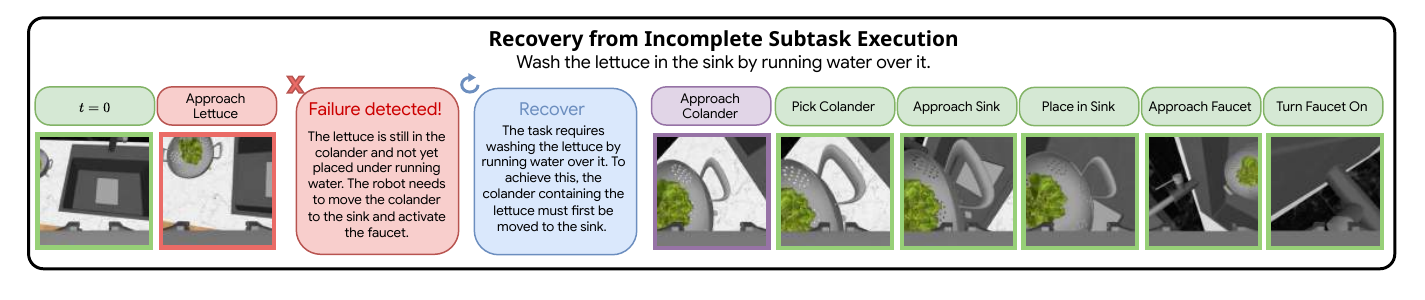}
        \caption{Correcting incomplete execution by transferring the lettuce-filled colander to the sink and activating the faucet.}
        \label{fig:incomplete_execution}
    \end{subfigure}

    \caption{
        \textbf{Qualitative analysis of failure-conditioned recovery.}
        Three representative manipulation failures and their corresponding recovery sequences. RePlan-VLA uses visual progress monitoring to identify discrepancies between intended and observed execution, generates diagnostic feedback, and revises the remaining plan. The examples illustrate recovery from an incorrect object grasp, restoration of a missing manipulation prerequisite, and completion of previously omitted actions. All three executions ultimately succeed.
    }
    \label{fig:qualitative_recovery}
    \vspace{-5mm}
\end{figure*}

We conduct ablation studies to assess five key components of RePlan-VLA: visual grounding, instruction verification, replanning context, planning strategies, and progress estimation. 
Each experiment modifies a single component of our framework while keeping the rest unchanged to isolate its contribution. 
We report success rate (SR) on RoboCasa365 composite tasks, with results presented in~\cref{tab:component_ablations}.

\begin{table}[t]
\centering
\footnotesize
\caption{
Component ablations of RePlan-VLA on RoboCasa365 composite tasks. SR is the task success rate (\%). Each ablation modifies only one component of the framework. Higher is better.
}
\label{tab:component_ablations}

\setlength{\tabcolsep}{5pt}
\renewcommand{\arraystretch}{0.95}

\begin{tabular}{lrrr}
\toprule
& \multicolumn{3}{c}{Success Rate (\%)} \\
Configuration
    & Seen & Unseen & Avg. \\

\midrule

Visual Grounding & & & \\
\ \ \ \ SAM3 Segmentation Masks & $46.8$ & $34.7$ & $40.7$ \\
\ \ \ \ Grounding DINO Bounding Boxes & $43.8$ & $31.2$ & $37.5$ \\
\ \ \ \ Object Labels Only & $37.5$ & $31.3$ & $34.4$ \\
\ \ \ \ Without Object Annotations & $37.2$ & $37.5$ & $37.4$ \\
\midrule

Instruction Verification & & & \\
\ \ \ \ Without Instruction Verifier & $37.5$ & $31.4$ & $34.45$ \\
\midrule

Replanning Context & & & \\
\ \ \ \ Without Completed Subtask History & $37.5$ & $31.3$ & $34.4$ \\
\ \ \ \ Without Failed Subtask & $18.7$ & $25.0$ & $21.9$ \\
\ \ \ \ Without Failure Explanation & $37.5$ & $18.8$ & $28.1$ \\
\midrule

Planning Strategy & & & \\
\ \ \ \ Zero-Shot Prompting & $37.5$ & $18.8$ & $28.1$ \\
\ \ \ \ Without Object-Specific Prompts & $50.0$ & $18.7$ & $34.4$ \\
\ \ \ \ Without Scene Observations & $37.5$ & $37.5$ & $37.5$ \\
\midrule

Progress Estimation & & & \\
\ \ \ \ With Zero-Shot VLM & $43.8$ & $31.3$ & $37.6$ \\
\ \ \ \ With Global Estimation~\cite{zhang2026recurrent} & $18.7$ & $18.8$ & $18.8$ \\

\midrule

\textbf{Full RePlan-VLA} & $\mathbf{68.8}$ & $\mathbf{56.3}$ & $\mathbf{62.5}$ \\

\bottomrule
\end{tabular}
\vspace{-5mm}
\end{table}

\noindent\textbf{Effect of Visual Grounding Strategies}.
We investigate how different object detection and scene annotation strategies affect the performance by influencing the planner's ability to generate scene-grounded subtasks. %
As shown in~\cref{tab:component_ablations}, SAM3 bounding boxes achieve the highest average success rate (62.5\%), outperforming SAM3 segmentation masks (46.9\%), Grounding DINO boxes (37.5\%), object labels alone (34.4\%), and unannotated observations (37.4\%). 
This highlights the importance of accurate object localization and consistent visual representations for scene-grounded planning. 
While providing finer object boundaries, segmentation masks may introduce visual ambiguity through inconsistent mask appearances across observations.
In contrast, labeled bounding boxes offer stable spatial cues and object identities. 
The limited effectiveness of object labels alone further suggests that identifying relevant objects without localizing them is insufficient for reliable planning.

\noindent\textbf{Effect of Instruction Verification}.
We examine the effect of validating generated subtask instructions against the underlying VLA's supported language forms on the performance of our approach. %
As shown in~\cref{tab:component_ablations}, removing the verification reduces average success from 62.5\% to 34.5\%, highlighting the importance of aligning generated subtasks with the underlying VLA's learned instruction patterns. 
Although the decomposer may generate semantically valid instructions, unfamiliar language formulations can lead to unreliable execution by the VLA policy. 
By identifying incompatible instructions and prompting their reformulation, the verifier bridges high-level planning and the VLA's execution capabilities. 
The consistent improvement across both splits demonstrates that successful planning requires not only selecting appropriate subtasks, but also expressing them in language the policy can reliably execute.

\noindent\textbf{Ablation Over Replanning Context}.
We investigate the contributions of the execution history, failure explanations, and subtask instruction towards generating contextually appropriate recovery plans.
As shown in~\cref{tab:component_ablations}, omitting the failed subtask causes the largest decline in average success (62.5\% to 21.9\%), highlighting the importance of identifying where execution failed. 
Removing completed-subtask history reduces success to 34.4\%, emphasizing its role in preserving achieved goals and avoiding redundant actions. 
Notably, excluding failure explanations particularly affects unseen tasks, reducing success from 56.3\% to 18.8\%. 
Together, these findings demonstrate the complementary importance of completed-subtask history, failure localization, and diagnostic feedback for effective recovery planning.

\noindent\textbf{Different Planning Strategies}.
We evaluate the effects of few-shot prompting, object-specific guidance, visual scene context, and task-conditioned object selection on plan quality and downstream execution.
As shown in~\cref{tab:component_ablations}, removing few-shot demonstrations, object-specific prompts, or scene observations substantially reduces task success. 
Notably, zero-shot prompting and removing object-specific guidance decrease unseen-task success from 56.3\% to 18.8\% and 18.7\%, respectively, highlighting the importance of structured examples and manipulation-specific knowledge for unfamiliar task compositions. 
Without scene observations, success drops to 37.5\% on both splits, underscoring the role of visual context in reasoning about object states and prerequisites. 
These findings demonstrate the complementary benefits of task-specific guidance and visual grounding for reliable long-horizon planning.

\noindent\textbf{Alternative Progress Estimation Methods}.
We compare our fine-tuned progress estimator with zero-shot VLM-based monitoring and global task-progress estimation~\cite{zhang2026recurrent} to assess the benefits of supervised primitive-level progress estimation.
As illustrated in~\cref{tab:component_ablations}, our fine-tuned primitive-level progress estimator achieves 62.5\% average success, outperforming zero-shot VLM monitoring (37.6\%) and R$^2$VLM-based global estimation~\cite{zhang2026recurrent} (18.8\%). 
While zero-shot reasoning lacks explicit manipulation-progress supervision, global estimation tracks overall task advancement rather than individual primitive completion, where substantial progress toward the overall goal does not necessarily imply successful completion of the current manipulation primitive; such discrepancies can lead to premature transitions or delayed corrective interventions. %
In contrast, our estimator provides progress feedback aligned with the execution granularity. 
These findings highlight the importance of task-specific supervision and fine-grained progress monitoring for reliable long-horizon manipulation.

\subsection{Qualitative Failure Recovery Analysis}
\label{sec:qualitative}

We qualitatively investigate how RePlan-VLA handles execution failures during long-horizon manipulation tasks. 
We examine representative video rollouts and their corresponding reasoning traces, analyzing how the progress monitor identifies execution discrepancies and how the decomposer adapts subsequent subtasks. 
We observe several distinct failure-recovery behaviors. First, our approach handles incorrect object manipulation, as shown in~\cref{fig:incorrect_object,fig:teaser}, where the robot grasps an unintended object, but the system recognizes the discrepancy and revises the plan to accommodate the current state. 
Second, it addresses missing manipulation prerequisites as illustrated in~\cref{fig:missing_prerequisite}, where repeated unsuccessful retrieval attempts lead to the introduction of an additional subtask. %
Third, it recovers from incomplete subtask execution as shown in~\cref{fig:incomplete_execution}, where the system identifies unfinished placements or interactions and redirects execution toward completing them before proceeding. 
Together, these examples illustrate how visual progress monitoring and adaptive task decomposition enable recovery from execution errors. 
Further analysis of recovery modes, along with a complete episode rollout example, is provided in the Appendix. 

\subsection{Real-World Franka Research 3 Deployment}
We evaluate RePlan-VLA on a Franka Research 3 robot performing $3$-task manipulation sequences following a CALVIN-style evaluation protocol. 
Using a fine-tuned and subsequently frozen $\pi_{0.5}$-DROID~\cite{intelligence2025pi_} policy, RePlan-VLA successfully completes long-horizon sequences of tasks, demonstrating its applicability to physical manipulation under real-world uncertainties. 
Detailed results and qualitative recovery rollouts are provided in the Appendix.

\subsection{Runtime and Replanning Analysis}
We evaluate RePlan-VLA's computational efficiency and replanning behavior on RoboCasa365, reporting component inference latencies, replanning frequency, and planner regeneration rates in~\cref{tab:runtime_analysis}. 
RePlan-VLA achieves an average end-to-end execution time of approximately $0.58$\,s per environment step, including policy inference, simulation, and framework overhead. 
Progress estimation ($\sim$2.35\,s/call) and object detection ($\sim$0.53\,s/call) incur additional monitoring costs, while replanning occurs $1.34$ times per episode on average. 
The primary overhead arises from planner inference, particularly when instruction verification failures trigger re-querying in $25.3\%$ of requests, compared with just $1.3\%$ for malformed outputs. 
Although verification is inexpensive ($\sim$0.22\,s/call), rejected instructions necessitate additional planner calls. 
Overall, our results indicate that execution-aware monitoring and adaptive replanning introduce manageable overhead in terms of planning time.

\begin{table}[t]
\centering
\footnotesize
\caption{
Runtime efficiency and replanning statistics of RePlan-VLA
on RoboCasa365. Latencies are reported in seconds per call.
Replanning frequency is averaged per episode, while planner
regeneration rates are percentages of decomposition requests.
}
\label{tab:runtime_analysis}

\setlength{\tabcolsep}{10pt}
\renewcommand{\arraystretch}{1.05}

\begin{tabular}{lrr}
\toprule
Metric & Seen & Unseen \\
\midrule
\multicolumn{3}{l}{Runtime Efficiency} \\
\ \ \ \ Progress Estimation (s/call) & 2.31 & 2.39 \\
\ \ \ \ Re-planning (s/call) & 3.56 & 2.69 \\
\ \ \ \ Instruction Verification (s/call) & 0.23 & 0.21 \\
\ \ \ \ Object Detection (s/call) & 0.57 & 0.51 \\
\midrule
\multicolumn{3}{l}{Re-planning Behavior} \\
\ \ \ \ Re-plans per Episode & 1.62 & 1.06 \\
\midrule
\multicolumn{3}{l}{Planner Reliability} \\
\ \ \ \ Malformed Output Re-query (\%) & 0.00 & 3.03 \\
\ \ \ \ Verification Failure Re-query (\%) & 30.95 & 18.18 \\
\midrule
\textbf{Average Time per Step (s/step)} & $\mathbf{0.580}$ & $\mathbf{0.587}$ \\

\bottomrule
\end{tabular}
\vspace{-5mm}
\end{table}

%% file: sec/5_Conclusion.tex
\section{Conclusion and Future Work}
\label{sec:conclusion}
We present RePlan-VLA, a VLA-agnostic framework that improves long-horizon manipulation through visual progress estimation, failure-aware feedback, and adaptive replanning while preserving the capabilities of pretrained VLA policies. 
Although its effectiveness depends on reliable visual grounding and the underlying policy's short-horizon manipulation capabilities, and recovery may be limited by irreversible execution failures, RePlan-VLA provides a flexible approach to execution-aware long-horizon planning without modifying the underlying VLA. 
Future work will explore skill-aware replanning that accounts for policy capabilities, learning from successful recoveries, and incorporating tactile sensing and other complementary modalities to improve progress estimation.